\pdfoutput=1

\documentclass[11pt]{article}
\usepackage[final]{EMNLP2022}
\usepackage[utf8]{inputenc}
\usepackage{amsfonts}
\usepackage{amsmath}
\usepackage{xspace}
\usepackage{ textcomp }
\usepackage{float}
\restylefloat{table}
\usepackage{array}
\usepackage{graphics}
\usepackage{graphicx}
\usepackage[frozencache,cachedir=.]{minted}
\newcommand\blfootnote[1]{%
  \begingroup
  \renewcommand\thefootnote{}\footnote{#1}%
  \addtocounter{footnote}{-1}%
  \endgroup
}
\usepackage{algorithm}
\usepackage{algpseudocode}
\usepackage{times}
\usepackage{latexsym}
\usepackage[T1]{fontenc}
\usepackage[utf8]{inputenc}
\usepackage{microtype}

\usepackage{inconsolata}

\title{DSFormer: Effective Compression of Text-Transformers by \\
Dense-Sparse Weight Factorization}

\author{Rahul Chand* \\
  Microsoft Research \\
  \texttt{chandrahul0320@gmail.com} \\\And
  Yashoteja Prabhu \\
  Microsoft Research \\
  \texttt{yprabhu@microsoft.com} \\\And
  Pratyush Kumar \\
  Microsoft Research \& IIT Madras\\
  \texttt{pratyush@iitm.ac.in} \\}

\begin{document}
\maketitle
\begin{abstract}

\blfootnote{*Work completed during time at Microsoft Research}

With the tremendous success of large transformer models in natural language understanding, down-sizing them for cost-effective deployments has become critical. Recent studies have explored the low-rank weight factorization techniques which are efficient to train, and apply out-of-the-box to any transformer architecture. Unfortunately, the low-rank assumption tends to be over-restrictive and hinders the expressiveness of the compressed model. 

This paper proposes, DSFormer, a simple alternative factorization scheme which expresses a target weight matrix as the product of a small dense and a semi-structured sparse matrix. The resulting approximation is more faithful to the weight distribution in transformers and therefore achieves a stronger efficiency-accuracy trade-off. Another concern with existing factorizers is their dependence on a task-unaware initialization step 
which degrades the accuracy of the resulting model. DSFormer addresses this issue through a novel Straight-Through Factorizer (STF) algorithm that jointly learns all the weight factorizations to directly maximize the final task accuracy. 

Extensive experiments on multiple natural language understanding benchmarks demonstrate that DSFormer obtains up to $40\%$ better compression than the state-of-the-art low-rank factorizers, leading semi-structured sparsity baselines and popular knowledge distillation approaches. Our approach is also  orthogonal to mainstream compressors and offers up to $50\%$ additional compression when added to popular distilled, layer-shared and quantized transformers. We empirically evaluate the benefits of STF over conventional optimization practices.
\end{abstract}

\section{Introduction}
Large pre-trained transformer models such as BERT~\citep{devlin19-bert} have achieved tremendous success in various natural language understanding (NLU) tasks including text classification \citep{wang18-glue}, question answering \citep{rajpurkar16-squad}, and summarization \citep{liu19-textsummarization}. Unfortunately, these large models also incur exorbitant serving costs when deployed in practical applications. In addition, they are unaffordable for resource-constrained settings such as inference on mobile or edge devices. This has inspired recent research on down-sizing these models without significant quality degradation.

The model compression approaches exploit the parameter redundancies present in a pre-trained model~\citep{frankle19-lotteryticket} to learn a smaller task-specific network with reduced working memory, storage, runtime or energy demands. A popular technique for transformer compression is to distil the knowledge of a pre-trained teacher network into a small student network by training from a large corpus~\citep{sanh19-distilbert,jiao20-tinybert}. The distillation approaches have been shown to achieve high compression ratios but require extravagant training and student architecture search which limit its utility and generality. The other mainstream approaches involve accelerating the attention dot products~\citep{kitaev20-reformer}, parameter sharing across network layers~\citep{lan20-albert} or quantizing individual weights of a network~\citep{bai21-binarybert}. These only reduce model size or inference time or are dependent on specialized hardware. 

\begin{figure}
\centering
\includegraphics[width=75mm]{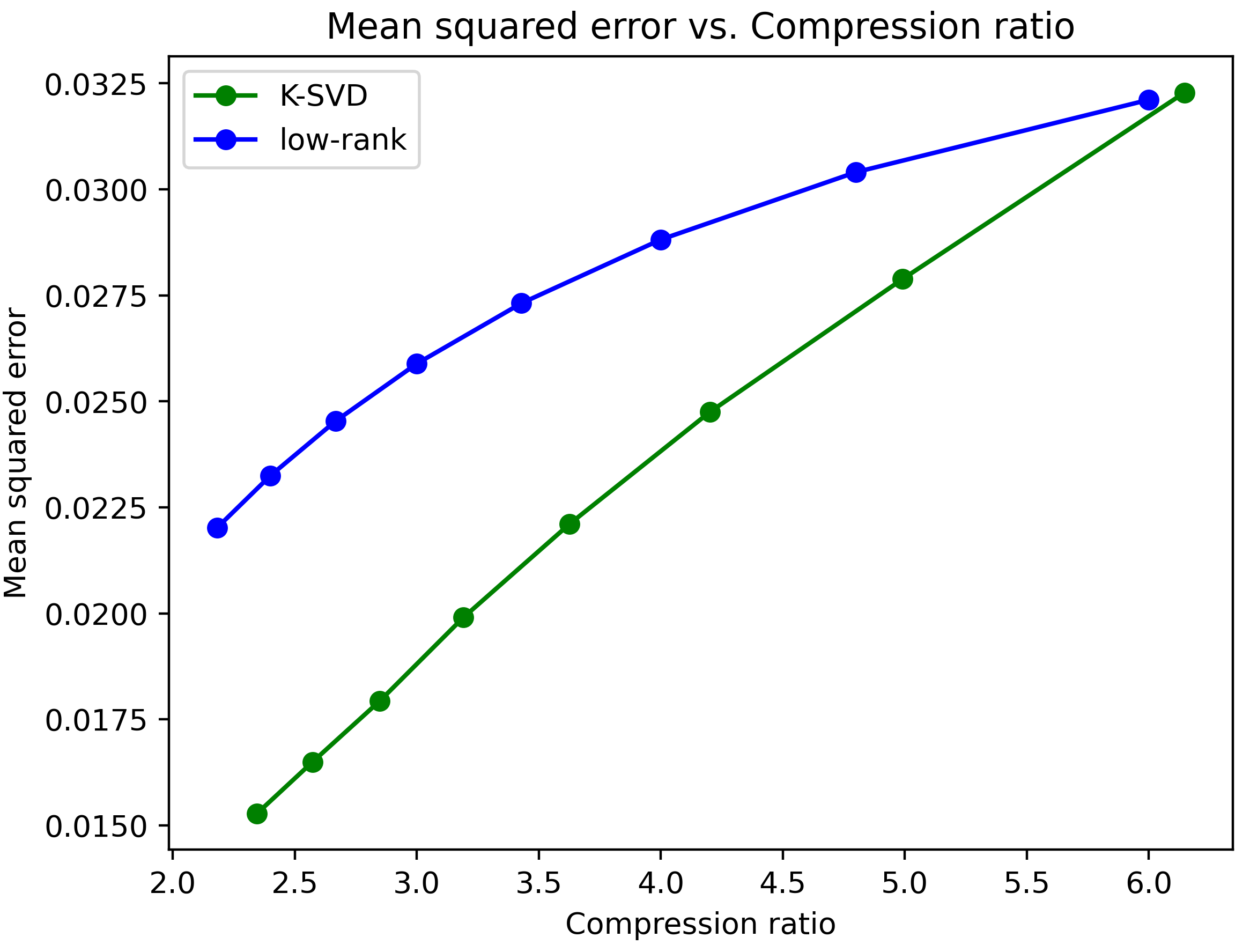}
\caption{Approximation errors due to low-rank factorization and dense-sparse factorization (using K-SVD algorithm). For similar compression ratios, the dense-sparse scheme results is strictly smaller errors. The reference weights are from the 6th layer of \texorpdfstring{BERT\textsubscript{BASE}}}
\label{fig:error_plt}
\end{figure}

Another line of recent work explores the use of lightweight low-rank factorization techniques to derive an efficient and generic compression scheme~\citep{chen21-drone,hsu21-fwsvd}. 
These approaches decompose the dense weight matrices, which account for a major portion of a transformer's model size, as product of two smaller matrices. This also accelerates the expensive dense matrix multiplication operations in the multi-head attention and feed-forward layers at inference time. Such factorization techniques are applicable to any transformer architecture and provides practical runtime benefits on commodity CPUs and GPUs. Notwithstanding these merits, the low-rank assumption does not typically hold for transformer weights and hence yields insufficient compression.

To address these limitations, we propose a simple alternative factorization scheme, named DSFormer, which utilizes a more-relaxed {\it locally low-rank} assumption. DSFormer approximates each weight vector as a linear combination of a small, but personalized, sub-set of basis vectors. Such an scheme is mathematically equivalent to a product of a small dense and a semi-structured sparse matrix. The resulting approximation is more accurate than existing low-rank alternatives (see Figure~\ref{fig:error_plt}) and yields up to $40\%$ higher compression for comparable final accuracies on standard NLU datasets.  Besides superior compression, the DSFormer scheme is also efficient and lends itself to hardware-friendly implementations owing to its useful semi-structured sparsity property. DSFormer approach is also generic and orthogonal to mainstream compression alternatives such as distillation, layer sharing and pruning, and is found to yield up to $50\%$ additional compression without hurting the prediction quality significantly.

For high compression rates, the weight factorization in DSFormer should preferably be learnt to directly optimize a task-specific objective. However, since factorization operations are expensive, the existing approaches resort to a task-agnostic single-shot factorization which is error-prone and degrades the final model accuracy. Instead, we explore more efficient alternatives to factorization which are amenable to task-aware optimization. In particular, we extend the popular straight-through estimator~\citep{bengio13-ste} trick to factorization operation to obtain a novel Straight-Through Factorizer (STF) optimizer which consistently the improves the model quality. Although STF is designed for DSFormer, its core ideas are applicable to any factorization scheme and deserves wider exploration in the future.

Our contributions are: (1) A powerful factorization scheme called DSFormer which offers a stronger efficiency-accuracy trade-off than existing alternatives; (2) A novel optimization technique called Straight-Through Factorizer (STF) which provides more accurate factorizations than existing training techniques; (3) Comprehensive comparative and ablative analysis on various NLU datasets to study the effectiveness of DSFormer and STF.

\section{Related Work}

Owing to its importance, several classes of techniques for transformer compression have been proposed in the literature. These include distillation~\citep{sanh19-distilbert,jiao20-tinybert,sun20-mobilebert}, quantization~\citep{zafrir19-q8bert,shen2020-qbert,bai21-binarybert}, pruning~\citep{frankle19-lotteryticket,guo19-proxpruning,gordon20-pruningeffect}, parameter-sharing~\citep{reid21-subformer,lan20-albert}, factorization~\citep{mao20-ladabert,chen21-drone,hsu21-fwsvd,chen18-groupreduce} etc. Among these, the low-rank factorization and the semi-structured pruning techniques are most relevant to this paper.

\textbf{Low-rank factorization}: These approaches attempt to compress the large dense weight matrices by SVD-based low-rank decomposition. DRONE~\citep{chen21-drone} first observed that vanilla low-rank approximations degrade the performance. It further, conjectured that the effective sub-space used by input data is smaller and proposed a {\it data-aware} factorization upgrade. FW-SVD~\citep{hsu21-fwsvd} further observed that all weight parameters are not equally important for accuracy on a given task and used the fisher-information criteria for incorporating the parameter importance into factorization step. These improvements succeed in providing slightly better compression, but nevertheless lose $>2\%$ accuracy points at larger compression rates of even 2x or more.

\textbf{Semi-structured pruning}: These approaches leverage the recent hardware innovations in accelerating semi-regular $N:M$ sparsity where each $N$ contiguous values in a weight matrix contain exactly $M$ non-zeros. For example, NVidia Ampere architecture~\citep{nvidia-4x2support} supports $4:2$ sparsity and provides $~2$x speed-ups on recent GPUs. Due to this, several approaches have been proposed to cleverly induce the $N:M$ sparsity into dense weight matrices for maximum performance. ASP~\citep{mishra21-asp} proposed training the dense network until convergence, using single-shot magnitude-based pruning to induce sparsity conformant to the NxM constraints, and then repeating the fine-tuning to recover the accuracy drop. SR-STE~\citep{zhou21-srste} used a sparse-refined straight-through estimator to learn the sparsity throughout the entire training, but resorted to costly pre-training for recovering accuracy. NxMTransformer~\citep{holmes21-nxmtransformer} proposed an efficient alternative based on ADMM algorithm for learning $N:M$ sparsity without need for pre-training. Our proposed dense-sparse factorization can be considered as a generalization of such $N:M$ techniques when the dense factor is assumed to be identity matrix. None of the $N:M$ optimization techniques are directly applicable to the general DSFormer setting since they do not support non-trivial factorization.

\begin{figure*}
\centering
\includegraphics[width=100mm]{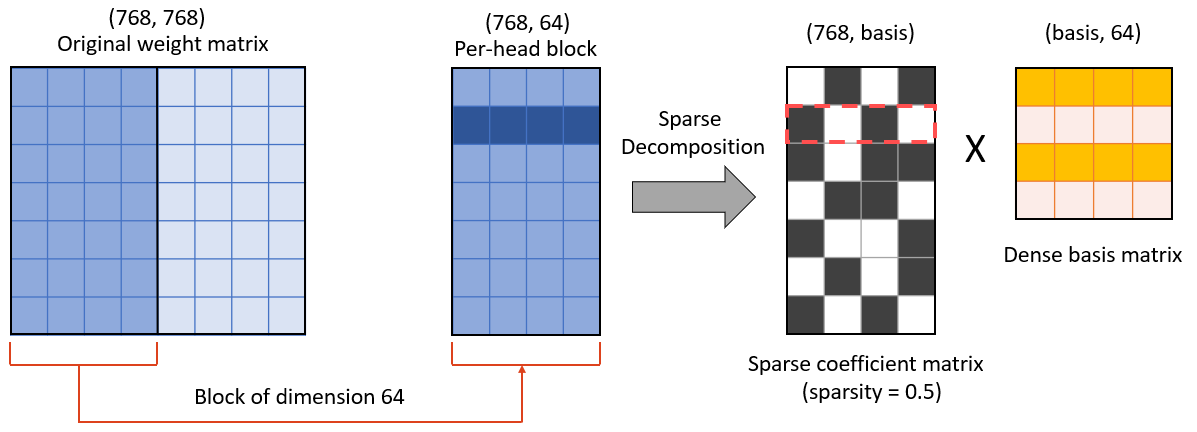}
\caption{Illustration of the decomposition scheme used in DSFormer. The original weight matrix is divided into blocks and each block is then represented using a sparse coefficient matrix and a dense basis matrix. The highlighted row in blue is composed of the coefficients in the red dotted row and the orange highlighted basis vectors.}
\label{fig:arch_fig}
\end{figure*}

\label{sec:ftfft}
\section{Preliminaries}

This section presents the generic architecture of a transformer and introduces the weight factorization framework for model compression.

\subsection{Transformer Architecture}
The impressive performance of transformers in language modeling has led to a proliferation of transformer model varieties with a few examples being BERT~\citep{devlin19-bert}, RoBERTa~\citep{liu19-roberta}, TinyBERT~\citep{jiao20-tinybert} and ALBERT~\citep{lan20-albert}. All these models share the same underlying architecture which comprises three components: an embedding layer, multiple blocks of self-attention and a classifier head.

The embedding layer performs a dictionary look-up of input token ids to obtain a sequence of embeddings. These embeddings are then successively refined by self-attention blocks which capture the long-range dependencies inherent in a text sequence. The final block's output embeddings are pooled and used by the classifier head for task-specific decisions such as whether the text contains $+ve$ or $-ve$ sentiment etc. Generally, the self-attention blocks account for a major portion of the model size ($78\%$ in BERT-base) and inference time. 

A self-attention block is further composed of two main sub-blocks: multi-head attention (MHA) and fully-connected feed-forward network (FFN). The MHA computation utilizes four set of dense weight matrices, namely, query ${\bf W}^q$, key ${\bf W}^k$, value ${\bf W}^v$ and output ${\bf W}^o$. For an input embedding sequence ${\bf X}_0$, the output of MHA is modeled by:

\begin{align}
    &{\bf S} = \text{softmax}(({\bf W}^{q}\cdot{\bf X}_0)^{\top}({\bf W}^{k}\cdot{\bf X}_0)) \\
    &{\bf X}_1 = {\bf W}^o\cdot( {\bf S}({\bf W}^{v}\cdot{\bf X}_0) )
\end{align}

where $\cdot$ indicates matrix-matrix product and $\sigma$ refers to layer-norm operator. The intermediate embedding matrix ${\bf X}_1$ is passed through FFN layer which operates with additional weight matrices ${\bf W}^{f_{1}},{\bf W}^{f_{2}}$ as ${\bf X}_2 = {\bf W}^{f_{2}}\cdot ReLU({\bf W}^{f_{1}}\cdot{\bf X}_1) $.

A typical transformer contains several self-attention blocks which increase the representative power of the learnt model. Apart from these, a transformer also involves several lightweight bias parameters and layer-norm operators for regularization which are usually resource-efficient and are hence ignored for the rest of the discussion. 

\subsection{Transformer Weight Factorization}

Transformers are high-capacity models of more than a 100 million parameters and making them more compact and affordable is essential to ensure their broader applicability. A major portion of these parameters are contained in the self-attention blocks, specifically among their $6$ large weight matrices ${\bf W}^{\{q,k,v,o,f_{1},f_{2}\}}_l$ each of which participates in a matrix-matrix operation with intermediate input representations ($l \in \{1\cdots L\}$ is the block number). The weight-factorization approaches aim to replace each such weight matrix by an equivalent  lower-capacity network while keeping the overall structure of the model in-tact.

Let $\mathcal{W} = \cup_{l=1}^L {\bf W}^{\{q,k,v,o,f_{1},f_{2}\}}_l$ be the set of all self-attention weights. For a ${\bf W}_{M\times N} \in \mathcal{W}$, its factorized representation and the associated dot-product with inputs are respectively:

\begin{align}
    &{\bf W}_{M\times N} \approx {\bf U}_{M\times K}\cdot{\bf V}_{K\times N} \\
    &{\bf W}\cdot{\bf X} \approx {\bf U}\cdot({\bf V} \cdot{\bf X})
\end{align}

In low-rank factorization approaches~\citep{chen21-drone,hsu21-fwsvd}, both ${\bf U},{\bf V}$ are dense matrices with the low-rank condition $K < \frac{MN}{M+N}$ necessary for non-trivial compression and speed-up. 

The factorization approaches largely adopt a three-staged training schedule for effective learning. The pre-trained model is first fine-tuned for a given task in the standard manner without any compression. This provides the best weight configurations which are then approximated through low-rank factorization. Finally, the factorized network is again fine-tuned in an end-to-end manner to recover the errors introduced by the approximation. We denote this schedule of double fine-tuning with a factorization by "FT-F-FT".

\section{Proposed Method: DSFormer}
This section presents the dense-sparse factorization scheme, discusses the compression gains that are offered by this approach, and proposes a novel optimizer for learning the factorization structure in a task-aware manner.

\subsection{Dense-Sparse Factorization}

The transformer weights in all its layers have been found to be geometrically well spread-out and occupy the full-dimensional space which maximizes its representation power. As a result, the search for low-rank structures has turned out to be mostly futile for transformers. We address the underlying issue by making a more relaxed {\it locally low-rank} assumption which provides a better approximation for the same compression ratio.

Figure~\ref{fig:arch_fig} illustrates the DS-factorization scheme. It adopts a block-wise factorization scheme where a weight matrix is divided into multiple thin blocks of width $B$ and each block is separately approximated by applying dense-sparse factorization. Consider a candidate weight matrix ${\bf W}_{M\times N} \in \mathcal{W}$ which can be expressed equivalently as a horizontal stack of blocks $[{\bf W}_1,\cdots,{\bf W}_{\frac{N}{B}}]$ where $N \% B = 0$ is assumed and ${\bf W}_i ~\forall i \in \{1,\cdots \frac{N}{B}\}$ is of dimension $M \times B$. The dense-sparse factorization expresses each $\bf{W}_i$ as:

\begin{align}
    &{\bf W}_i \approx {\bf S}_i \cdot {\bf D}_i \\
    &{\bf S}_i \in \{M \times K\}\\
    &\|{\bf S}_i[j,:]\|_0  = S ~~\forall j \in \{1,\cdots,M\}\\
    &{\bf D}_i \in \{K \times B\}
\end{align}

The inner dimension $K$ is chosen to satisfy $B < K \ll M$ where $K>B$ ensures that ${\bf D}_i$ can be of full-rank $B$ and therefore represent  the whole space of ${\bf W}_i$, and $K \ll M$ ensures compactness of the factorized form. Each row of ${\bf W}_i$ is approximated as a linear combination of exactly $S$ rows of ${\bf D}_i$ owing to the imposed sparsity-constraint on rows of ${\bf S}_i$. The value $S$ satisfies $S<B$ and hence, although ${\bf S}_i$ contains more elements than ${\bf W}_i$ itself, the non-zero values are far fewer and thereby promote compression. In summary, an ${\bf W}_i$ row is locally approximated by a small subset of nearby ${\bf D}_i$ rows. The total choice of such subsets, $\binom{K}{S}$, is huge even for a reasonably small $S$ value which boosts the expressive power of DS-factorization scheme. Note that low-rank factorization can be considered a special case of our approach when $S=K<B$.

A useful interpretation of DS-factorization can be obtained by a block-diagonal and a horizontal stacking of ${\bf D}_i$ and ${\bf S}_i$ factors, respectively, to obtain ${\bf S} \in \{M \times \frac{NK}{B}\}, {\bf D} \in \{\frac{NK}{B}\times N\}$ which satisfy ${\bf W} = {\bf S}\cdot{\bf D}$. The sparse ${\bf S}$ factor possesses exactly $S$ non-zeros in each contiguous range of $K$ row values. Such a sparsity is referred to as semi-structured sparsity in compression literature~\citep{holmes21-nxmtransformer} and oftens leads to efficient implementations on supported hardware.

The standard approach to recover such factors ${\bf D}_i,{\bf S}_i$ for a given ideal ${\bf W}_i$ is to solve a least-squares optimization problem:

\begin{align}
    \label{eqn:ksvd}
    &\min_{{\bf D},{\bf S}} \|{\bf W}_i - {\bf S}\cdot{\bf D}\|^2\\
    &\text{s.t.}~~\|{\bf S}_i[j,:]\|_0  = S ~~\forall j \in \{1,\cdots,M\}
\end{align}

Due to its sub-set constraints, such problems are known to be NP-hard and lack exact algorithmic solutions that are also efficient. We instead utilize a well-known heuristic solver called K-SVD for minimizing (\ref{eqn:ksvd}). A GPU-based parallel implementation of Batch K-SVD algorithm ~\citep{rubinstein08-ksvd} was used.

\begin{table*}[t]
  \centering
  
 \resizebox{15cm}{!}{
  \begin{tabular}{c|c|c c c c c c c c | c }
  \hline
    {} & CR & CoLA & SST-2 & MRPC & STS-B & MNLI-m & QNLI & RTE & QQP & Average\\
    \hline
    \texorpdfstring{BERT\textsubscript{BASE}} && 1x & 58.06 & 92.74 & 89.4 & 89.19 & 84.1 & 91.6 &  69.28 & 89.6 & 82.05\\
    \hline
    DSFormer  & \textbf{2x} & 58.16 & 93.09  & 90.15  & 89.24 & 83.83 & 91.0 & 71.88 & 87.69 & 83.13\\
    DSFormer  & \textbf{2.8x} & 57.8 & 92.11 & 90.08 & 88.67 & 83.42 & 90.68 & 70.94 & 86.75 & 81.85\\
    DSFormer  & \textbf{3.57x} & 49.9 & 91.87 & 88.19 & 87.63 & 82.04 & 89.62 & 69.28 & 86.41 & 79.80\\
    \hline
    DistilBERT & 2x & 51.3 & 91.3 & 87.5 & 86.9 & 82.2 & 89.2 & 59.9 & 88.5 &78.33\\
    TinyBERT-6 & 2x & 52.84 & 93 & 90.6 & 89.6 & 84.5 & 91.1 & 73.4 & 88.0 & 82.88\\
    DRONE \(\dagger\) & 1.5x & 53.2 & 90.8 & 88.0 & 87.8 & 82.6 & 89.3 & 71.5 & 90.1  &80.46\\
    FWSVD & 2x & 49.4 & 91.2 & 88.0 & 87.0 & 83.0 & 89.5 & - & 87.6 & -\\
    NxMTransformer  & 1.88x & 55.3 & 92.3 & 90.8 & 89.3 & 82.3 & 90.4 & 68.6 & - &81.29\\
    ASP & 1.88x & 51.7 & 91.9 & 90.8 & 88.8 & 83.3 & 90.6 & 68.6 & - & 80.81 \\
    \hline
    
  \end{tabular}
  }
  \caption{Dev set results on GLUE benchmark. The metrics for these tasks can be found in the GLUE paper~\citep{wang18-glue}. The results show that DSFormer outperforms all competing baselines at 2x compression on the average GLUE score. For higher compression range of 2.8x it is withing 1\% point of \texorpdfstring{BERT\textsubscript{BASE}}{}  and DSFormer consistently outperforms competing baselines like DistilBERT, DRONE and ASP on the benchmark while being 50\%, 86\% and 40\% smalleracross all tasks. The performance remains competitive even in the more extreme 3.57x compression scheme. The last column provides the arithmetic average of accuracies across all tasks.  \(\dagger\)Denotes that the compression ratio is the average compression across all tasks. 
  }
  \label{tab:1}
\end{table*}

\subsection{Compression \& Speed-up Ratios}

The number of bytes for storing ${\bf D}_i,{\bf S}_i$ factors with full-precision are respectively $4KB$ and $4SM$ where the latter is restricted to non-zero count in ${\bf S}_i$. Following ~\citep{holmes21-nxmtransformer,mishra21-asp}, we also include additional $\frac{\log_2 K}{8}SM$ bytes for storing the non-zero indices of ${\bf S}_i$. In comparison, the original weight block ${\bf W}_i$ required $4MB$ bytes. For simplification, let's re-parameterize $K = \gamma M$ and $S = \delta B$ with $\gamma,\delta < 1$. Then the compression ratio is:

\begin{align}
    \label{eqn:cr}
    CR  &= \frac{KB + (1+\frac{\log_2 K}{32})SM}{MB} \nonumber\\
    &= \gamma + (1+\frac{\log_2 \gamma + \log_2 M}{32})\delta
\end{align}

Here $\gamma,\delta$ are the hyper-parameters that control the degree of compression with smaller values yielding more compact models. Note that $M$ is at most a few 1000s for most transformers. 

During inference, the matrix multiplication with an input embedding matrix ${\bf X}_{N \times L}$ is implemented through block-level product operations as follows:

\begin{align}
    \label{eqn:inference}
    {\bf W} \! \cdot \! {\bf X} &= \sum_{i=1}^{\frac{N}{B}} {\bf W}_i \! \cdot \! {\bf X}_i = \sum_{i=1}^{\frac{N}{B}} {\bf S}_i \! \cdot \! \big({\bf D}_i\cdot {\bf X}_i\big) 
\end{align}

where ${\bf X}_i$ refers to a block of $B$ contiguous rows of ${\bf X}$. Since each parameter in ${\bf D}_i,{\bf S}_i$ is used exactly once during multiplication with a column of ${\bf X}$, the improvement in flops is also approximately the same as in (\ref{eqn:cr}).

\subsection{Efficient Inference on GPU \& CPU}

For practical usefulness, it is desirable that a compression technique yields models which support efficient implementations on available hardware and thus provide realistic speed-ups. 

The inference step of DS-factorization (\ref{eqn:inference}) involves a dense and a semi-structured sparse matrix product operations. The dense mat-muls are well-known to be efficient and capable of taking advantage of vector processing architectures of modern CPUs and GPUs. Additionally, the recent hardware innovation trends, such as sparse-tensor cores of NVidia Ampere architecture~\citep{nvidia-4x2support}, have been improving the hardware-support for exploiting the semi-structured sparsity.

A less-studied aspect of semi-structured sparsity is its advantages on standard CPUs with memory hierarchies. Typically, the sparse matrix products are notorious for having a lower computational intensity which measures the number of computations performed for every memory access~\citep{williams07-compintensity}. Since the memory accesses tend to be several orders slower than an arithmetic operation, larger computational intensity is preferable for faster inferences. Interestingly, the block structure and regular sparsity of DS-factorization create predictable and local data access patterns during mat-muls which allow more computations for every memory access.

\begin{figure}
\centering
\includegraphics[width=80mm]{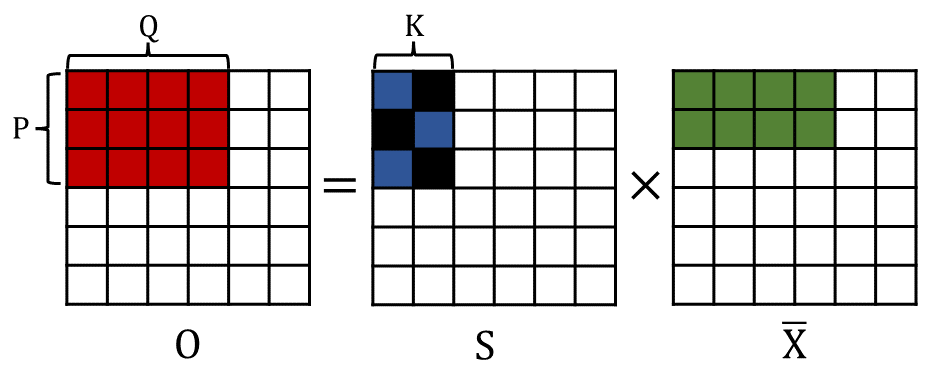}
\caption{Implementation of a block-wise semi-structured sparse mat-mul for effective cache utilization}
\label{fig:sparsematmul}
\end{figure}

To understand this, consider the operations ${\bf O} = {\bf W}\cdot{\bf X} = {\bf S}\cdot({\bf D}\cdot{\bf X})$ on a simplified two-tier memory hierarchy with a large-but-slow RAM and a faster-but-smaller cache with a limited storage of $C$ bytes. The first dense-product ${\bf \bar{X}} = {\bf D}\cdot{\bf X}$ involves well-studied and efficient dense mat-muls, therefore we mainly focus on the more challenging ${\bf O} = {\bf S}\cdot{\bf \bar{X}}$ mat-mul. As illustrated in Figure~\ref{fig:sparsematmul}, for efficient cache utilization, we follow a block-structured implementation where the complete mat-mul is broken down into smaller mat-muls over sub-matrices of ${\bf O},{\bf S},{\bf \bar{X}}$. In one step, the blocks of sizes ${P \times Q}$ from ${\bf O}$, ${P \times K}$ from ${\bf S}$ and ${K \times Q}$ from ${\bf \bar{X}}$ are loaded into cache. The amount of memory accessed from RAM is

\begin{align}
    \label{eqn:cacheload}
    PQ + PS + KQ = C 
\end{align}

 where cache is assumed to be fully utilized. The number of computations possible after cache transfer are $PQS$ products after ignoring a few  read/write operations with faster cache. Therefore, the computational intensity is:

\begin{align}
    CI  &= \frac{PQS}{PQ+PS+KQ} = \frac{PQS}{C}
\end{align}

Intuitively, if $P,Q,S,K$ are comparable, then $PQS>C$ is expected thus providing non-trivial speed-ups. More generally, for a given choice of $S,K,C$, we can maximize the above $CI$ by using the linear relation between $P,Q$ in (\ref{eqn:cacheload}) to obtain closed-form solutions for optimal $P,Q$.

\subsection{Training with Straight-Through Factorizer}

The objective of task-specific compression is to learn a compact model which offers high accuracy on a given task. Therefore, it is desirable to jointly learn all weight factorizations in order to directly optimize the task objective. Unfortunately, the DS-factorization involves discrete sparsity constraints which prevents a naive end-to-end optimization using the standard gradient-based approaches.

This challenge is side-stepped by existing approaches with a  three-staged FT-F-FT schedule elaborated in Section~\ref{sec:ftfft}. FT-F-FT involves fixing the sparsity-structure through a single-shot K-SVD factorization and then learning only the non-zero values using traditional gradient optimization. Although this enables efficient learning, we observe such a schedule to be sub-optimal due to two reasons: (1) The sparsity-structure is fixed task-agnostically and is hence not optimized for metric of importance; (2) The single-shot approximate K-SVD solution can be located far-away from an ideal minimum and hence fail to fully recover its error by gradient descent on a non-convex manifold.

Another popular solution to such discrete optimization problems involves using Straight-Through Estimation (STE)~\citep{bengio13-ste} trick. STE performs a fresh discretization in each forward pass which allows its approximation to be refined over the course of training. At the same time, STE avoids the ill-defined differentiation over discrete variables by entirely ignoring the discretization module in the backward passes. The STE heuristic has been widely adopted for popular discrete approximations like  Quantization~\citep{ying19-stequantization}, Pruning~\citep{srinivas17-stepruning} etc. Unfortunately, less attention has been paid to STE for factorization.

We study the use of STE trick for learning DS-factorizations effectively. A straight-forward idea would be to use K-SVD for discretization in every forward pass; however such a naive approach will be extremely slow since K-SVD factorization takes several iterations to converge. Moreover, performing a full K-SVD in each forward pass is wasteful; the underlying weight $\bf{W}$ changes only slightly from one step to the next and therefore its factors $\bf{D},\bf{S}$ are also expected to change gradually.

These observations inspire an efficient STE adaptation, referred to as Straight-Through Factorizer (STF), that gradually refines its factors over the course of training. The core objective of STF is to solve the least-squares problem in (\ref{eqn:ksvd}) which gives the desired factors $\bf{D},\bf{S}$. Unlike K-SVD, STF gradually optimizes over the continuous-valued $\bf{D}$ by performing just a single gradient-update step with respect to (\ref{eqn:ksvd})  per backward pass. As the target ${\bf W}$ itself is continuously updating, $\bf{D}$ slowly converges towards the final value of ${\bf W}$. The discrete $\bf{S}$ factor is, however, learnt freshly in each forward pass by fixing ${\bf W},{\bf D}$ and minimizing (\ref{eqn:ksvd}). Such  minimization problems are well-known in sparse-coding literature and we borrow a standard solver named Orthogonal Matching Pursuit (OMP)~\citep{zhang11-omp}. We use a GPU-based fast version of OMP which allows efficient training. The pseudo-code for an STF pass is provided in Appendix.

To leverage STF, we modify the FT-F-FT training schedule in Section~\ref{sec:ftfft} to FT-F-STF. We retain the original two stages of vanilla fine-tuning followed by factorization. However, in the third stage, we continue to refine the ${\bf W}$ values by STF. As shown in Section~\ref{sec:exp}, FT-F-STF consistently achieves better accuracies than traditional FT-F-FT schedule, especially when compression demands are higher. The STF stage is affordable and completes in at most 2 hours on largest of our datasets.

\section{Experiments}
\label{sec:exp}

In this section, we evaluate our approach and show its effectiveness in compressing Transformer networks on various NLU benchmarks.

\subsection{Baselines}
The DSFormer approach is compared to state-of-the-art factorization baselines DRONE~\citep{chen21-drone} and FWSVD~\citep{hsu21-fwsvd}, state-of-the-art semi structured sparsity approaches, NxMTransformer~\citep{holmes21-nxmtransformer} and ASP~\citep{mishra21-asp} and popular distillation baselines, TinyBERT~\citep{jiao20-tinybert} and DistilBERT~\citep{sanh19-distilbert}. For reliable comparison, the author-provided numbers have been used for all baselines.

\subsection{Model Settings}
The compression in DSFormer depends on two hyper-parameters, the ratio between number of basis vectors and output dimension \(\gamma\) and ratio between coefficient matrix sparsity and head size \(\delta\). We can select \(\gamma\) and \(\delta\) to control the trade-off between model size and accuracy depending on the requirements. For our experiments we use two configurations, \(\gamma\), \(\delta\) \(=(1/4, 3/16)\) and \(\gamma\), \(\delta\) \(=(1/4, 1/4)\) which were found to work well across all evaluation tasks and provide a compression of 2.8x and 3.57x respectively with respect to \texorpdfstring{BERT\textsubscript{BASE}}{}, making them suitable for comparison to competing baselines. Other hyper-parameters, i.e.\ number of layers, embedding size and number of heads remain same as  \texorpdfstring{BERT\textsubscript{BASE}}{}. 

\begin{figure}
\centering
\includegraphics[width=76mm]{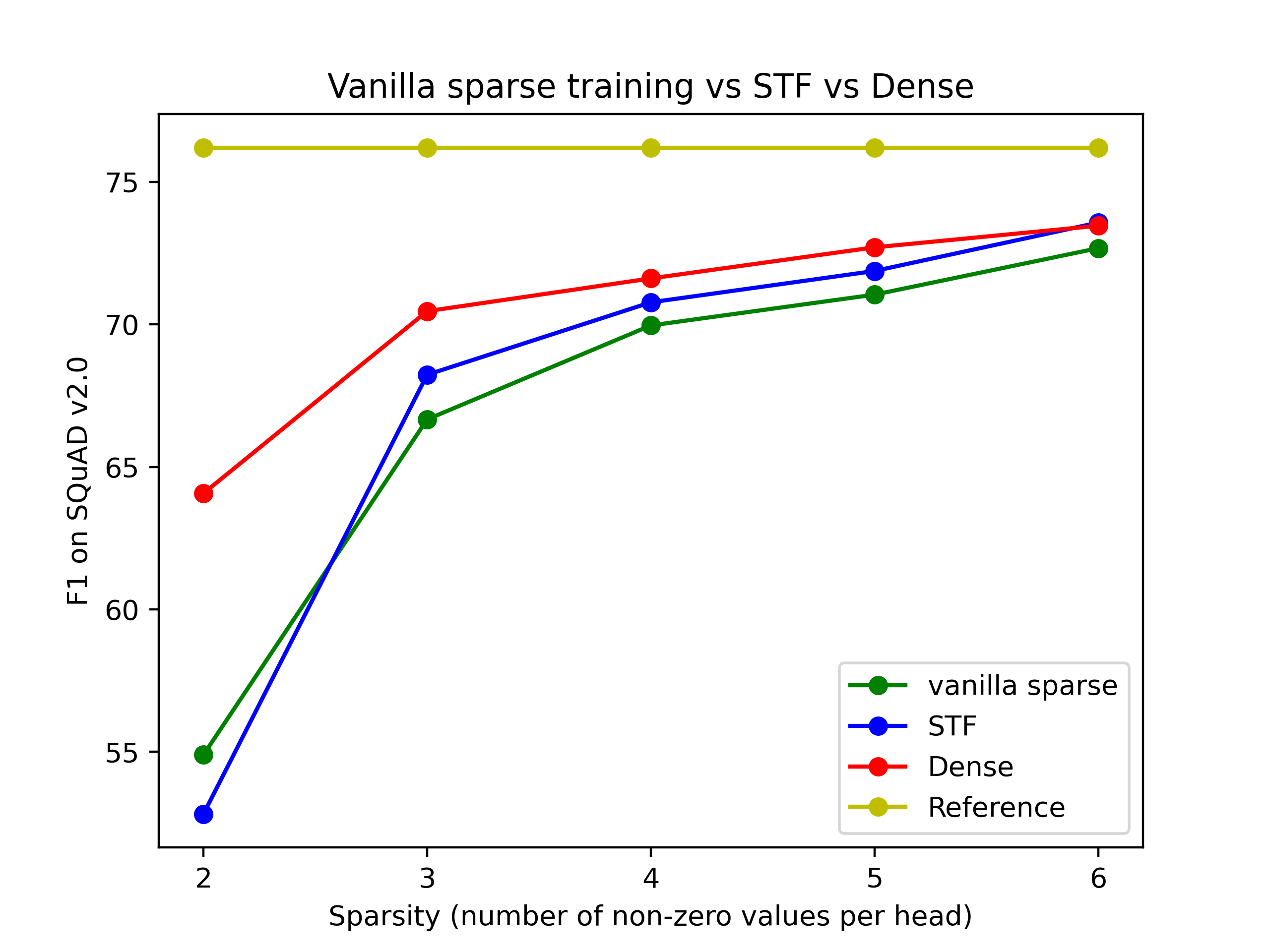}
\caption{Comparison between FT-F-STF, vanilla sparse finetuning (FT-F-FT), vanilla dense finetuning for low sparsity values (also plotted is BERT base reference F1)}
\label{fig:an_stf_plt}
\end{figure}

\subsection{Training Settings}
As noted before, our compression is focussed on the dense matrices in self-attention blocks which account for most of the transformer size. The remaining parameters such as layernorms, biases, classifier head and embedding layer are optimized in their native forms. All parameters are initialized from pre-trained \texorpdfstring{BERT\textsubscript{BASE}}{}. 

For each task, DSFormer is fine-tuned using FT-F-STF schedule and with Adam optimizer~\citep{https://doi.org/10.48550/arxiv.1412.6980}. A grid search is performed over different batch sizes $(8/16/24/32)$, learning rates ($5e^{-5}$ to $e^{-4}$) and number of epochs $(1-5)$ to determine the best hyper-parameters. DSFormer is trained on 8 32GB V100 GPUs, fine-tuning takes between 4 minutes (RTE, CoLA, MRPC) to 2 hours (SQuAD v2.0). 

\subsection{Results on GLUE}
The General Language Understanding Evaluation (GLUE) benchmark~\citep{wang18-glue} is a collection of NLP tasks for evaluating natural language understanding systems. We report scores on the development set for each task, the results are summarized in Table ~\ref{tab:1} (F1 scores are reported for MRPC, Spearman correlations are reported for STS-B and accuracy scores are reported for other tasks). 

The experiment results show that DSFormer scores within $1\%$ of the reference \texorpdfstring{BERT\textsubscript{BASE}}{} numbers while being both 2.8x smaller and faster. DSFormer outperforms DistilBERT in accuracy despite a $40\%$ greater compression. In comparison to TinyBERT-6, DSFormer's accuracy is $1\%$ worse; however it is unfair to directly compare with TinyBERT since the latter utilizes expensive task-specific data augmentation while finetuning. When compared against DRONE and FWSVD, the approaches that are more similar in spirit to DSFormer, it outperforms both on all GLUE tasks except one while being 86\% and 40\% smaller respectively. DSFormer is a generalization of NxM semi-structured sparsity approaches and significantly out-performs both ASP and NxMTransformer in compression while also improving accuracy by $1\%$.

\subsection{Results on SQuAD}
We also evaluate DSFormer on question answering (QA) datasets: SQuAD v1.1~\citep{rajpurkar16-squad} and SQuAD v2.0~\citep{rajpurkar-etal-2018-know}. For fair comparison, we also include DSFormer numbers for compression ratio of 2x. The hyper-parameters used for this configuration are \(\gamma\)\(= (5/36, 1/6, 13/96, 1/6)\) for QKV, O, FFN-1 and FFN-2 respectively, while \(\delta\) \(= 9/32\) is same for all components. As shown in Table~\ref{tab:3}, DSFormer outperforms all competing baselines on SQuAD v1.0 by 0.5\% point. For the tougher SQuAD v2.0 task, DSFormer again outperforms all baselines except TinyBERT-6. Even when compression ratio is increased to 3.57x, DSFormer still retains 97\% of \texorpdfstring{BERT\textsubscript{BASE}}{} performance. 

\subsection{Effect of optimization technique on DSFormer performance}
We conjecture that the conventional FT-F-FT fine-tuning is sub-optimal due to the initial approximation being far from the optimal value. As improvements, the STF technique and the FT-F-STF schedule were proposed in Section 4.4. To validate the hypothesis, we conduct experiments for small sparsity values of 4 to 6 where the single-shot factorization is expected to be more erroneous; the results are summarized in Figure~\ref{fig:an_stf_plt}. As can be observed, DSFormer outperforms vanilla sparse training by a large margin. As expected, the gap reduces with higher sparsity ranges where initial approximation error is smaller.

\subsection{Orthogonality to existing techniques}

Transformer compression area contains many existing techniques and it is desirable for a new technique be orthogonal to existing approaches. In this section, we conduct experiments to study how effectively DSFormer can be applied on top of well-known layer-sharing techniques, knowledge distillation and quantization approaches. 

For each experiment, we take an already compressed and finetuned model, ALBERT for parameter sharing, DistilBERT for knowledge distillation, and Q8BERT for quantization. The hyper-parameters used in this experiment are same as that in Section 5.5. The results are summarized in Table ~\ref{tab:2}. As observed, DSFormer can provide 2X additional compression with minimal additional drop in accuracy over most of these baselines. In the case of ALBERT with SQuAD, the performance drop is more than 2\% point. This is expected since ALBERT is already highly compressed (12x) and SQuAD is one of the tougher datasets. 

\section{Conclusions}

Transformer compression is presently a critical topic of research with numerous down-stream applications. The recently proposed low-rank factorization approaches to compressing transformers have the significant advantages of simplicity, generality and hardware-friendliness, but suffer from insufficient compression. This paper studied the limitations of existing approaches and prescribed novel techniques in the form of a dense-sparse scheme for compression and a straight-through factorizer algorithm for optimizing such a scheme. The resulting DSFormer can result in up to $40\%$ better compression than leading baselines while also being orthogonal to mainstream compression alternatives. The ideas in STF are general to any factorization scheme and can inspire more explorations in the future.

\begin{table}
  \resizebox{220pt}{!}{
  \centering
  \begin{tabular}{l | c | c c c c c c c c c}
  \hline
    {} & CR & SQuAD & SST-2 & QNLI & MRPC & MNLI & RTE & STSB & CoLA & QQP\\
    \hline 
    ALBERT & 12x & 89.3 & 90.3 & 89.94 & 90.4 & 83.8 & 70.7 & 90.3 & 57.7 & 87.5 \\
    \hspace{0.45cm} + FT-F-FT & \textbf{24x} & 86.5 & 90.2 & 89.02 & 89.4 & 82.7 & 65.7 & 87.6 & 43.5 & 86.8 \\
    DistilBERT & 2x & 86.9 & 91.3 & 89.2 & 87.5 & 82.2 & 59 & 86.6 & 57.0 & 88.5\\
    \hspace{0.45cm} + FT-F-FT\(\dagger\) & \textbf{4x} & 84.5 & 89.2 & 87.5 & 88.3 & 81.6 & 61 & 86.4 & 55.5 & 86.7\\
    Q8BERT & 4x & 87.74 & 92.24 & 90.62 & 89.6 & \textbf{N/A} & 68.8 & 89.0 & 58.5 & 88.0\\
    \hspace{0.45cm} + FT-F-FT & \textbf{8x} & 87.37 & 91.63 & 90.13 & 88.4 & 83.3 & 67.0 & 87.6 & 56.2 & 87.8\\
    \hline

  \end{tabular}
  }
  \caption{Result on applying DSFormer with distillation, 8-bit Quantization and parameter sharing on SQuAD v1.0 and selected GLUE tasks. The drop in performance is minimal, indicating that DSFormer is largely orthogonal to existing compression techniques. \(\dagger\) indicates that task specific distillation was not performed after factorization.
  }
  \label{tab:2}
\end{table}

\section*{Limitations}

This paper has two main limitations. First, the end-to-end wall-clock speed-ups due to DSFormer on actual hardware have not been demonstrated and are considered as beyond the scope of this paper. Second, DSFormer is task-specific and therefore requires learning separate model for every task; multi-task extensions can be considered for future work. 

\section*{Ethics Statement}

The following paper proposes a new compression scheme for transformer networks that are widely used across the world for natural language tasks and therefore it inherits the same ethical concerns that come with an NLP system. Similarly since the model is benchmarked on top of two standard datasets, GLUE and SQuAD it also implicitly contains biases and limitations that exist in the underlying datasets.

\section*{Acknowledgements}

\bibliography{custom}
\bibliographystyle{acl_natbib}

\appendix
\label{sec:appendix}
\centering
\section{Appendix}

\begin{table*}[!t]
  \centering
  \begin{tabular}{c |c | c c c c}
  \hline
    Model & \textbf{CR} & \multicolumn{2}{c}{SQuAD v1.1} & \multicolumn{2}{c}{SQuAD v2.0}\\
    {} & {} & \textbf{EM} & \textbf{F1} & \textbf{EM} & \textbf{F1} \\
    \hline 
    \texorpdfstring{BERT\textsubscript{BASE}} && 1x & 80.8 & 88.5 & 72.73 & 76.2 \\
    \hline
    DSFormer & 2x & \textbf{79.91} & \textbf{88.01} & \underline{72.21} & \underline{75.7}\\
    DSFormer & 2.8x & 79.6 & 87.42 & 71.88 & 75.15\\
    DSFormer & 3.57x & 78.7 & 86.9 & 71.3 & 74.5\\
    \hline
    DistilBERT & 2x & 77.7 & 86.9 & 66.0 & 69.5\\
    TinyBERT-6 & 2x & \underline{79.7} & 87.5 & \textbf{74.7} & \textbf{77.7} \\
    ASP & 1.88x & 79.17 & \underline{87.52} & 71.9 & 75.6\\
    \hline
  \end{tabular}
  \caption{Dev results of baselines and DSFormer on SQuAD datasests.  DSFormer beats all competing baselines on SQuAD v1.1 and is withing 0.5\% point of \texorpdfstring{BERT\textsubscript{BASE}}{} while being 50\% smaller. Similarly DSFormer also outperforms all competing baselines except TinyBERT-6 on SQuAD v2.0, which as noted previously can be an unfair comparison due to the latter utilizing data augmentation while finetuning. Even for the more extreme compression of 3.57x DSFormer remains competitive. 
  }
  \label{tab:3}
\end{table*}  

\begin{table*}[!t]
    \centering
  \begin{tabular}{c | c c c}
  \hline
    Dataset &  \multicolumn{3}{c}{Head size} \\
    {} & 32 & 64 & 128\\
    \hline 
    SQuAD v1.1 & 79.6/87.9 & 79.9/88.0 & \textbf{80.26/88.1}\\
    SQuAD v2.0 & \textbf{72.6/76.0} & 72.2/75.7 & 72.1/75.3\\
    SST-2 & 91.8 & \textbf{93.1} & 91.6\\
    QNLI & 90.78 & \textbf{91.0} & 90.18\\
    MRPC & 89.66 & \textbf{90.16} & 88.1\\
    MNLI & \textbf{83.93} & 83.83 & 82.72\\
    RTE & 71.52 & \textbf{71.88} & 70.625\\
    STSB & \textbf{89.54} & 89.24 & 89.09\\
    CoLA & \textbf{59.14} & 58.16 & 55.0\\
    QQP & 86.40 & \textbf{87.69} & 87.57\\
    \hline
  \end{tabular}
  
  \caption{Ablation study on the effect of head-size on DSFormer performance on all GLUE tasks \& SQuAD v1.1. All models in this Table have the same compression ratio of \textbf{2x}.
  }
  \label{tab:4}
\end{table*}

\begin{table*}[!t]
    \centering
  
  \begin{tabular}{c | c | c c c c}
  \hline
    Model & \textbf{CR} & Pre-training? & Data-augmentation? & Add. pre-processing & Training Time (m) \\
    \hline 
    
    DSFormer & 2.8x & \texttimes & \texttimes & \checkmark & - \\
    DSFormer & 3.57x & \texttimes & \texttimes & \checkmark & - \\
    \hline
    DistilBERT & 2x & \checkmark & \texttimes & \texttimes & - \\
    TinyBERT-6 & 2x & \checkmark & \checkmark & \texttimes & - \\
    DRONE & 1.5x & \texttimes & \checkmark & \checkmark & - \\
    FWSVD & 2x & \texttimes & \checkmark & \checkmark & - \\
    NxMTransformer & 1.88x & \texttimes & \texttimes & \texttimes & 120 \\
    ASP & 1.88x & \texttimes & \texttimes & \checkmark & - \\
    \hline
  \end{tabular}

  \caption{Training Time comparison b/w DSFormer and competing baselines. The Training time is time taken to train 
  }
  \label{tab:5}
\end{table*}

\begin{algorithm}
\label{code:stf}
\caption{Pseudo code for STF}
\label{code:stf}
\begin{minted}
[
fontsize=\footnotesize,
]
{python}

class STF(autograd.Function):
    
    @staticmethod
    def forward(ctx, W, D, S):
        return torch.mm(S, D)
    
    @staticmethod
    def backward(ctx, g):
        gradW = g
        gradD = torch.mm(S.t(), torch.mm(S,D) - W)
        gradS = None
        
        return gradW, gradD, gradS
\end{minted}
\end{algorithm}

\end{document}